\documentclass[11pt,a4paper]{article}
\usepackage[hyperref]{naaclhlt2018}
\usepackage{times}
\usepackage{latexsym}
\usepackage{graphicx}
\usepackage{enumitem}
\usepackage{url}

\usepackage{amsmath}

\usepackage[utf8x]{inputenc}
\usepackage{subcaption}

\aclfinalcopy 

\title{Duluth at SemEval-2019 Task 6:\\Lexical Approaches to Identify and Categorize Offensive Tweets}

\author{Ted Pedersen \\
Department of Computer Science \\
University of Minnesota\\
Duluth, MN 55812 USA \\
{\tt tpederse@d.umn.edu}}

\date{}

\begin{document}
\maketitle
\begin{abstract}
This paper describes the Duluth systems that participated
in SemEval--2019 Task 6, Identifying and Categorizing
Offensive Language in Social Media (OffensEval). 
For the most part these
systems took traditional Machine Learning
approaches that built classifiers from lexical features found 
in manually labeled training data. However, our most successful system for
classifying a tweet as offensive (or not) was a 
rule-based black--list approach, and we 
also experimented with combining the 
training data from two different but related SemEval tasks. 
Our best systems in each of the three OffensEval tasks 
placed in the middle of the 
comparative evaluation, ranking 57\textsuperscript{th} of 103 in task A, 
39\textsuperscript{th} of 75  in task B, and 
44\textsuperscript{th} of 65 in task C.
\end{abstract}

\section{Introduction}

Social media is notorious for providing a platform
for offensive,
toxic, and hateful speech. There is a pressing need for
NLP tools that can identify and moderate 
this type of content. 

The OffensEval task \cite{OffensEval} focuses on 
identifying offensive language
in tweets, and determining if specific individuals or
groups are being targeted. Our approach was to rely
on traditional Machine Learning methods as implemented by
Scikit \cite{Scikit-learn} to build classifiers from
manually labeled examples of offensive tweets. Our models
relied on lexical features, including
character ngrams, words, and word ngrams. We also included
a dictionary based black--list approach, and experimented
with combining training data from two related yet different
SemEval-2019 tasks.

Offensive language is an umbrella term that covers hate speech, cyber-bullying,
abusive language, profanity, and so on. Recognizing offensive language is an important first step in dealing with different kinds of problematic text. Language that is offensive may
simply violate community standards regarding the use of profanity,
but in other cases may cross over to become abusive, threatening, or dangerous. Detecting such
language has proven to be a challenging problem, at least in part because it remains difficult
to make distinctions between the casual and even friendly use of profanity versus more 
serious forms of offensive language \cite{FortunaN18,SchmidtW17}.

\section{Experimental Data}

OffensEval is made up of three tasks that were carried out
in stages during January 2019. Task A is to classify a tweet as being 
offensive (OFF) or not (NOT). Task B takes the offensive tweets
from task A and decides if they are targeted insults (TIN) or not (UNT). Task C looks at the targeted
insults from task B and classifies them as being directed against
an individual  (IND), group (GRP) or other entity (OTH). 

Task A provides 13,240 training tweets, of which 
8,840 (66.7\%) were not offensive (NOT). Task B is made up of the
4,400 training tweets that were offensive (OFF), where 
3,876 (88.1\%) of these are targeted insults (TIN). Task C includes
the targeted insults from task B, of which 
2,407 (62.1\%) were targeted against 
individuals (IND) and 1,074 (27.7\%) were against groups (GRP).
Additional details about the task data can be found in \cite{OLID}

The distribution of classes in the evaluation data was similar. 
Task A has 860 evaluation tweets of which 620 (72\%) were not offensive. 
Task B includes 240 offensive evaluation tweets, where 213 (89\%) were targeted insults. These made up the evaluation data for task C, where 100 (47\%) were against individuals, and 78 (37\%) were against groups.

The amount of training data is modest, particularly for
tasks B and C. In addition, the classes in Task B are quite skewed. Given these factors
we decided to rely on traditional Machine Learning
techniques, since these have the potential to perform well 
even with limited training data. 

\section{System Descriptions}

We created three systems for each of the three tasks. Preprocessing was very simple : tweets were converted to lower case and then tokenized into character ngrams, words, or word ngrams.

In task A we relied on unigram models. However, for tasks B and C we focused on character ngrams in order to try and find regularities in the smaller amounts of training data available for these tasks. 

\subsection{Task A}

The goal of Task A was to classify a tweet as offensive (OFF) or not (NOT). We relied on the
default settings for vectorization and Machine Learning as provided in Scikit, except as noted below. This results in 
tokenization based on space separated strings where punctuation
and other non-alphanumeric characters are discarded. 

A-Sub1 is a Logistic Regression classifier that weighs 
unigram features using tf-idf. 

A-Sub2 is the same as A-Sub1 except that the training
data is augmented with the training data from Semeval-2019
Task 5 HatEval \cite{hateval2019}, a task that identifies hate speech. 
It provides 22,240 training examples where 14,057 (65.2\%) are not 
hate speech. We made the obviously incorrect assumption that tweets that aren't 
hate speech would also not be
offensive. We had hoped that doubling the amount of training data
would improve our performance despite our flawed assumption (although it did not).

A-Sub3 is a very simple rule based on a black--list created by merging 
the following sources: 

\begin{itemize}[noitemsep]
\item words or terms used in offensive tweets five or more times in the OffensEval
training data, 
\item words or terms used in hateful tweets five or more times in the HatEval training data, and
\item black--lists found via web search said to be used by WordPress, Google, Facebook, and Youtube\footnote{https://www.freewebheaders.com/}.
\end{itemize}

Any word or term that appears in two or more of these lists is selected, 
leading to a master black--list of 563 words. Any tweet that includes any of
these words is labeled as offensive (OFF).

\subsection{Task B}

Task B seeks to identify if an offensive tweet is a targeted insult (TIN) or not (UNT). All three systems relied on character-based ngrams and the default Scikit settings for vectorization and the specific learning method. 

B-Sub1 learns 
a random forest classifier, and B-Sub2 learns a decision tree. 
Both represent features as 3 alphanumeric character sequences.

B-Sub3 learns 
a linear Support Vector Machine from the training data. Features
are represented as 3 non-space character sequences. We used non-space characters in
this case in order to capture special characters that would be discarded by B-Sub1 and B-Sub2. We were particularly interested in retaining \# (hashtags) and @ (user ids). 

\subsection{Task C}

All tweets in task C are targeted insults (TIN). The goal is to identify if the
target is an individual (IND), group (GRP), or some other entity (OTH). 

All of these systems used the default settings from Scikit 
for vectorization and Machine Learning, except as noted below. 

C-Sub1 learns
a multinomial Naive Bayesian classifier from the training data.
Features are simple unigrams made up of alpa-numeric characters. During development we noticed that
Naive Bayes tended to find more balanced distributions of classes 
than our other approaches. 

C-Sub2 learns
a decision tree classifier from the training data.
Features are 3 character sequences. During
development we observed that this was the only 
method that assigned tweets to
the other (OTH) category.

C-Sub3 learns
a logistic regression classifier from the training data. Features
are word ngrams made up of sequences of 1, 2 and 3 words that occur in more than 10
tweets in the training data. 

\section{Experimental Results}

The official rankings in OffensEval were based on macro--averaged F1, and accuracy was also reported. The performance of individual classes was  measured by Precision, Recall, and the $F1$ score.

The results of the Duluth Systems are summarized 
in Table \ref{table:results}. X-TOP is
the 1\textsuperscript{st} ranked system in each task. X-Baseline
assigns each
test tweet to the most frequent class in the training data.

\begin{table}
\centering
\begin{tabular}{lcc} 
System & F1-Macro & Accuracy\\
\hline
A-TOP & .83 &  \\
A-Sub3 & .73 & .80  \\
A-Sub2 & .69 & .80 \\
A-Sub1 & .68 & .80 \\
A-Baseline & .42 & .72 \\
\hline
B-TOP & .76 & \\
B-Sub1 & .60 & .88 \\
B-Sub2 & .57 & .82 \\
B-Sub3 & .52 & .77 \\
B-Baseline & .47 & .89 \\
\hline
C-TOP & .66 & \\
C-Sub1 & .48 & .67  \\
C-Sub3 & .45 & .62  \\
C-Sub2 & .42 & .53  \\
C-Baseline & .21 & .47 \\
\end{tabular}
\caption{Duluth OffensEval Results}
\label{table:results}
\end{table}

Next, we'll examine the results from
each task in more detail. In the confusion matrices provided, 
the distribution of gold answers (ground truth) is shown on the rows, and the system predictions are on the columns. 

\subsection{Task A}

Task A asks whether a tweet is offensive (or not). 
It had the largest amount of training data (13,240 examples), of which 33\% were considered offensive (OFF) and 67\% were not (NOT). In Table \ref{table:asub} and the discussion that follows a true positive is a tweet that
is known by ground truth to be not offensive (NOT) and that is predicted to be not offensive (NOT). A true negative is a tweet
that is known to be offensive (OFF) and is predicted to be offensive (OFF).

\subsection{Confusion Matrix Analysis}

A-Sub3 had a modest advantage over the other two systems. A-Sub3 was a simple rule-based black--list approach, while A-Sub1 and A-Sub2 used Machine Learning. All three systems scored identical accuracy (80\%), but in looking at their confusion matrices some interesting differences emerge. 

Table \ref{table:asub} shows that the rule based method A-Sub3 has a much smaller number of false negatives (112 versus 160 and 152). It also has a larger number of true negatives (128 versus 80 and 88). Overall  the rule based system finds more tweets offensive (190) than the Machine Learning methods (91 and 106). This happens because our rule based system only needs to find a single  occurrence of one of our 563 black--listed terms to consider a tweet offensive, no doubt leading to many non-offensive tweets being considered offensive (62 versus 11 and 18).

The only difference between A-Sub1 and A-Sub2 was that A-Sub2 had approximately double the number of training tweets. The extra tweets were from the SemEval-2019 hate speech task (HatEval). We hypothesized that more training data might help improve the results of a logistic regression classifier (which was used for both A-Sub1 and A-Sub2). After increasing the training data, A-Sub2 is able to classify exactly one more tweet correctly (690 versus 689). We were somewhat surprised at this very limited effect, although the HateEval corpus is focused on a particular domain of hate speech where the targets are women and immigrants. This does not appear to have matched well with the OffensEval training data.  

\begin{table}
\centering
\begin{tabular}{lrrr|rrr}
\hline
     &   NOT &   OFF & & P & R & F1 \\
 NOT &   609 &    11 & 620 & .79 & .98 & .88 \\
 OFF &   160 &    80 & 240 & .88 & .33 & .48 \\
     &   769 &    91 & 860 & .82 & .80 & .77 \\ \\
\multicolumn{7}{c}{A-Sub 1: Logistic Regression} \\ \\
\hline
     &   NOT &   OFF & & P & R & F1 \\
 NOT &   602 &    18 & 620 & .80 & .97 & .88\\
 OFF &   152 &    88 & 240 & .83 & .37 & .51\\
     &   754 &   106 & 860 & .81 & .80 & .77  \\ \\
\multicolumn{7}{c}{A-Sub 2: Logistic Reg + HatEval} \\ \\
\hline
     &   NOT &   OFF & & P & R & F1 \\
 NOT &   558 &    62 & 620 & .83 & .90 & .87\\
 OFF &   112 &   128 & 240 & .67 & .53 & .60 \\
     &   670 &   190 & 860 & .79 & .80 & .79 \\ \\
\multicolumn{7}{c}{A-Sub3 : Rule Based, black--list} \\ \\
\hline
\end{tabular}
\caption{Task A Confusion Matrices}
\label{table:asub}
\end{table}

\subsubsection{Feature Analysis}

Table \ref{table:asub1features} shows the top 30 most heavily weighted features according to the A-Sub1 logistic regression classifier (which was trained on 13,240 instances). 
We will have the convention of upper casing features 
indicative of an offensive tweet and lower casing not offensive features.
There are some stark differences between these feature sets, where the offensive 
ones are for the most part profanity and insults.

\begin{table}[t]
\centering
\begin{tabular}{l|l}
\multicolumn{1}{c}{NOT offensive} & 
\multicolumn{1}{c}{OFFensive} \\ \hline
     {\it user} &      {\bf SHIT} \\
    {\it antifa} &      {\bf FUCK}\\
      {\it url} &     {\bf BITCH}\\
      {\it best} &     {\bf STUPID}\\
     {\it thank} &     {\bf    ASS}\\
{\it conservatives} &   {\bf FUCKING}\\
       new &      {\bf IDIOT}\\
 {\it beautiful} &      {\bf LIAR}\\
      here & {\bf DISGUSTING}\\
    brexit &      {\bf SUCKS}\\
    {\it thanks} &      {\bf SICK}\\
     {\it love} &      {\bf CRAP}\\
       she &     {\bf RACIST}\\
    {\it   day} &      {\bf DUMB}\\
   awesome &    {\bf FASCIST}\\
  adorable &      NIGGA\\
      safe &     FUCKED\\
    voting &      {\bf CRAZY}\\
     funny &   {\bf IGNORANT}\\
     {\it stand} &       FOOL\\
   {\it justice} &     COWARD\\
      idea &     IDIOTS\\
     there &       SUCK\\
     right &       {\bf KILL}\\
      join &      PUSSY\\
      well &       UGLY\\
   amazing &      WORST\\
   {\it twitter} &       DAMN\\
   welcome &   BULLSHIT\\
    trying &    ASSHOLE\\
\end{tabular}
\caption{Task A Feature Analysis - A-Sub1}
\label{table:asub1features}
\end{table}

In Table \ref{table:asub2features} we show the top 30 weighted
features in A-Sub2, a logistic regression classifier trained on the combination of OffensEval and HatEval data. Terms relating to hatred of women and immigrants abound, and include numerous hash tags (recall that our tokenization only used alphanumerics so \# are omitted).  

\begin{table}[t]
\centering
\begin{tabular}{l|l}
\multicolumn{1}{c}{NOT offensive} & 
\multicolumn{1}{c}{OFFensive} \\ \hline
     https &     {\bf BITCH} \\
        co & BUILDTHATWALL \\
 immigrant &      {\bf SHIT} \\
       men &  WOMENSUCK \\
    {\it antifa} &       {\bf FUCK} \\
       {\it url} &        {\bf ASS} \\
      {\it user} &    ILLEGAL \\
       ram &    BITCHES \\
     {\it thank} &     {\bf SUCKS} \\
      {\it best} &     NODACA \\
       new & BUILDTHEWALL \\
{\it conservatives} &     {\bf STUPID} \\
      when &       {\bf LIAR} \\
       son &      {\bf IDIOT} \\
       you & {\bf DISGUSTING} \\
     {\it stand} &      WHORE \\
 {\it beautiful} &    {\bf FUCKING} \\
      kunt &        HOE \\
    {\it  love} &       SUCK \\
   {\it justice} &   ILLEGALS \\
  facebook &       {\bf SICK} \\
        ho &    {\bf FASCIST} \\
   tonight &      {\bf CRAP} \\
    {\it thanks} &   {\bf IGNORANT} \\
 wondering &      THESE \\
       {\it day} &     {\bf RACIST} \\
   accused &      {\bf  KILL} \\
    brexit &      {\bf CRAZY} \\
     alone &       {\bf DUMB} \\
   {\it twitter} &      WHITE \\
\end{tabular}
\caption{Task A Feature Analysis - A-Sub2}
\label{table:asub2features}
\end{table}

In Tables \ref{table:asub1features} and \ref{table:asub2features} we bold face the
18 features that were shared between A-Sub1 and A-Sub2. 
This gives us some insight
into the impact of merging the OffensEval and HatEval
training data. Some generic offensive features remain in Table
\ref{table:asub2features} but are strongly augmented by
HatEval features that are oriented against women and 
immigrants. 

The 13 shared terms that were indicative of the not
offensive class are shown in italics. Some features are 
what we'd expect for non-offensive tweets : {\it love, beautiful,
thanks, thank, justice} and {\it best}. Others are more artifacts of the
data, {\it user} is an anonymized twitter id and {\it url} 
is an anonymized web site. Others are less clearly not offensive, and seem related to political conversation : {\it antifa, conservatives,} and {\it brexit}. 

However, there are some inconsistencies to note. In Table \ref{table:asub1features} NIGGA is not 
necessarily an offensive term and points to the need for annotators to have
subtle understandings of culture \cite{WaseemTB2018}. In Table \ref{table:asub2features} {\it kunt} is a deliberate misspelling meant to 
disguise intent (c.f. \cite{GrondahlPKCA2018}).

Table \ref{table:asub3features} shows the top 30 terms from our black--list system A-Sub3 that proved to be most discriminating in identifying an offensive tweet. Recall that A-Sub3 had the highest F1-Macro score of our task A systems.  
The first column shows a simple ratio of the number of times a feature is used in an offensive tweet (OFF in 3\textsuperscript{rd} column)
versus a not offensive one (NOT in 4\textsuperscript{th} column). The most discriminating feature BITCH
occurred in 61 offensive tweets and in 0 that were not offensive.

\begin{table}[t]
\centering
\begin{tabular}{rlrl}
\multicolumn{1}{c}{Ratio} & 
\multicolumn{1}{c}{Feature} & 
\multicolumn{1}{c}{OFF} & 
\multicolumn{1}{c}{NOT} \\ \hline
61.0 &      BITCH &    61 &     0 \\
17.5 &      IDIOT &    35 &     2 \\
14.0 &    ASSHOLE &    14 &     0 \\
10.6 &       FUCK &   106 &    10 \\
10.2 &     STUPID &    92 &     9 \\
10.0 &       DICK &    10 &     1 \\
10.0 &    BITCHES &    10 &     0 \\
 9.0 &       SHIT &   278 &    31 \\
 9.0 &     RAPIST &     9 &     1 \\
 7.3 &     FUCKED &    22 &     3 \\
 6.3 &    FUCKING &    82 &    13 \\
 5.8 &      SUCKS &    35 &     6 \\
 5.5 &       CRAP &    33 &     6 \\
 5.3 &     IDIOTS &    16 &     3 \\
 5.0 &       SCUM &    10 &     2 \\
 5.0 &      MORON &    10 &     2 \\
 4.9 &        ASS &   108 &    22 \\
 4.8 &   IGNORANT &    19 &     4 \\
 4.5 &      LOSER &     9 &     2 \\
 4.3 &     SHITTY &    13 &     3 \\
 4.2 &       BUTT &    17 &     4 \\
 4.0 &       UGLY &    12 &     3 \\
 3.8 &       DUMB &    23 &     6 \\
 3.2 &      PUSSY &    13 &     4 \\
 3.2 &      NIGGA &    16 &     5 \\
 3.0 &       PORN &     9 &     3 \\
 2.9 &       HELL &    38 &    13 \\
 2.9 &   BULLSHIT &    23 &     8 \\
 2.6 &       SUCK &    21 &     8 \\
 2.5 &       KILL &    32 &    13 \\
\end{tabular}
\caption{Task A Feature Analysis - A-Sub3}
\label{table:asub3features}
\end{table}

\subsection{Task B}

Task B includes 4,400 training tweets, all of which are judged by ground truth to be offensive. This is a fairly modest amount of training data, particularly given how noisy tweets tend to be. As a result we shifted from using unigrams as features (as in Task A) and moved to the use of character ngrams, in the hopes of identifying patterns that may not exist at the unigram level. 

The data in Task B is also the most skewed of all the tasks. Nearly 90\% of the tweets belonged to the class of targeted insult (TIN). Our three Task B systems used different Machine Learning classifiers, and all tended to produce very skewed results, where most tweets were judged to be targeted insults (TIN). This is clearly illustrated in Table \ref{table:bsub}, which shows that the random forest classifier (B-Sub1) was better in terms of Precision and Recall for TIN, whereas all three classifiers struggled with the UNT class. 

\begin{table}
\centering
\begin{tabular}{lrrr|rrr}
\hline
     &   TIN &   UNT & & P & R & F1 \\
 TIN &   206 &     7 & 213 & .90 & .97 & .93 \\
 UNT &    22 &     5 & 27 & .42 & .19 & .26 \\
     &   228 &    12 & 240 & .85 & .88 & .86 \\ \\
\multicolumn{7}{c}{B-Sub1 : Random Forest} \\ \\
\hline
     &   TIN &   UNT & & P & R & F1 \\
 TIN &   188 &    25 & 213 & .90 & .88 & .89 \\
 UNT &    20 &     7 & 27 & .21 & .26 & .24 \\
     &   208 & 7    32 & 240 & .83 & .81 & .82 \\ \\
\multicolumn{7}{c}{B-Sub2 : Decision Tree} \\ \\
\hline
     &   TIN &   UNT & & P & R & F1 \\
 TIN &   179 &    34 & 213 & .90 & .84 & .87\\
 UNT &    21 &     6 & 27 & .15 & .22 & .18 \\
     &   200 &    40 & 240 & .81 & .77 & .79 \\ \\
\multicolumn{7}{c}{B-Sub3 : Linear SVM} \\ \\
\hline
\end{tabular}
\caption{Task B Duluth Systems}
\label{table:bsub}
\end{table}

\subsection{Task C}

Task C had an even smaller amount of training data (3,876 instances). Given a targeted insult, systems were asked to decide if the target an individual (IND), group (GRP) or other (OTHER). These appear as I, G, and O in Table \ref{table:csub}. 
The Other class is very sparse, and C-Sub1 and C-Sub3 did very poorly on it. However, C-Sub2 (a decision tree) had slightly more success. C-Sub1 and C-Sub2 rely on character ngrams, while C-Sub3 uses word unigrams, bigrams, and trigrams as features. 

\begin{table}
\centering
\begin{tabular}{lrrrr|rrr}
\hline
     &   G &   I &   O & & P & R & F1 \\
 G &    53 &    25 &     0 & 78 & .70 & .68 & .68 \\
 I &     9 &    90 &     1 & 100 & .66 & .90 & .76 \\
 O &    14 &    21 &     0 & 35 & .00 & .00 & .00 \\ 
     &    76 &    136 &    1 & 213 & .57 & .67 & .61 \\ \\
\multicolumn{8}{c}{C-Sub1 : Multinomial Naive Bayes} \\ \\
\hline
     &   G &   I &   O & & P & R & F1 \\
 G &    39 &    27 &    12 & 78 & .51 & .50 & .50\\
 I &    21 &    70 &     9 & 100 & .63 & .70 & .66 \\
 O &    17 &    15 &     3 & 35 & .13 & .09 & .10 \\
     &    77 &    112 &   24 & 213 & .50 & .53 & .51 \\ \\
\multicolumn{8}{c}{C-Sub2 : Decision Tree} \\ \\
\hline
     &   G &   I &   O & & P & R & F1 \\
 G &    48 &    25 &     5 & 78 & .61 & .62 & .61 \\
 I &    13 &    85 &     2 & 100 & .67 & .85 & .75 \\
 O &    18 &    17 &     0& 35 & .00 & .00 & .00 \\
     &    79 &    127 & 7 & 213 & .54 & .62 & .58 \\ \\
\multicolumn{8}{c}{C-Sub3 : Logistic Regression} \\ \\
\end{tabular}
\caption{Task C Duluth Systems}
\label{table:csub}
\end{table}

\section{Qualitative Review of Training Data}

Finally, we qualitatively studied some of the training data for task A and saw
that there is potential for some noise in the gold standard labeling. We
found various tweets labeled as offensive that seemed innocuous:


\begin{itemize}[noitemsep]
\item She should ask a few native Americans what their take on this is.
\item gun control! That is all these kids are asking for!
\item Tbh these days i just don't like people in general i
just don't connect with people these days just a annoyance..
\item Dont believe the hype.
\item Ouch!
\item Then your gonna get bitten
\item there is no need to be like That
\end{itemize}

We also found tweets labeled as not offensive despite the presence of insults
and profanity. 


\begin{itemize}[noitemsep]
\item Ppl who say I'm not racist are racist. You Are A Racist. Repeat after me
\item I'M SO FUCKING READY
\item Great news! Old moonbeam Just went into a coma!
\item No fucking way he said this!
\item Yep Antifa are literally Hitler.
\item Any updates re ending your blatant \#racism as \#Windrush \#Grenfell proves you are
\end{itemize}

The annotation guidelines from the OffensEval organizers 
seem relatively clear in stating that all profanity should
be considered offensive, although an annotator may intuitively wish to make a more nuanced
distinction. Resolving
these kinds of inconsistencies seems important since the data from task A is also used
for task B and C, and so there is a danger of unintended downstream impacts. 

\section{Conclusion}

Offensive language
can take many forms, and some words are offensive in one context but not another. As
we observed, profanity was often very indicative of offensive language, but of course
can be used in much more casual and friendly contexts. This quickly exposes the limits
of black--listing, since once a word is on a black--list it use will most likely always
be considered offensive. Identifying targeted targeted individuals or organizations
using lexical features and Machine Learning was extremely difficult, particularly 
given the small amounts of training data. Incorporating the use of syntactic analysis
and named entity recognition seem necessary to make progress.

We also encountered the challenging impact of domain differences in identifying offensive language.
Our attempt to (naively) increase the amount of available training data by combining
the OffensEval and HatEval data had no impact on our results, and our feature analysis
made it clear that the two corpora were different to the point of not really providing
much shared information that could be leveraged. That said, we intend to explore
more sophisticated approaches to transfer learning (e.g., \cite{KaranS2018,ParkF2017, WaseemTB2018}) since there are quite a few distinct corpora where various forms of hate speech have been annotated.

\bibliography{tdp}
\bibliographystyle{acl_natbib}

\end{document}